\documentclass[runningheads]{llncs}
\usepackage[T1]{fontenc}
\usepackage{graphicx}
\usepackage[hidelinks]{hyperref}
\usepackage[utf8]{inputenc} 
\usepackage[T1]{fontenc}    
\usepackage{booktabs}       
\usepackage{amsfonts}       
\usepackage{nicefrac}       
\usepackage{microtype}      
\usepackage{xcolor}

\usepackage{url}
\usepackage{graphicx}
\usepackage{color,xcolor}
\usepackage{amsmath,bm}
\newtheorem{thm}{Theorem}

\usepackage{multirow}
\usepackage{makecell}
\usepackage{algorithm}
\usepackage{algorithmic}
\usepackage{setspace}
\usepackage{xcolor}
\usepackage{amssymb} 
\usepackage{stfloats} 
\usepackage{wrapfig}
\usepackage{float}

\begin{document}
\title{On Leveraging Unlabeled Data for Concurrent Positive-Unlabeled Classification and Robust Generation}
\titlerunning{Concurrent Positive-Unlabeled Classification and Robust Generation}
%
\author{Bing Yu\inst{1}\thanks{Equal contribution, $\dagger$ Corresponding authors \email{zlin@pku.edu.cn, z.zhu@soton.ac.uk}} \and
Ke Sun\inst{2\star} \and
He Wang\inst{5} \and Zhouchen Lin\inst{2,3,4\dagger} \and Zhanxing Zhu\inst{6\dagger}}
%
%
\institute{School of Mathematical Sciences, Peking University \and
State Key Lab of General AI, School of Intelligence Science and Technology, Peking University \and
Institute for Artificial Intelligence, Peking University \and
Pazhou Laboratory (Huangpu), Guangzhou, Guangdong, China 
\and
Department of Computer Science, University College London \and
School of Electrical and Computer Science, University of Southampton
}	

\maketitle              
\begin{abstract}
The scarcity of class-labeled data is a ubiquitous bottleneck in many machine learning problems. While abundant unlabeled data typically exist and provide a potential solution, it is highly challenging to exploit them. In this paper, we address this problem by leveraging Positive-Unlabeled~(PU) classification and the conditional generation with extra unlabeled data \emph{simultaneously}. We present a novel training framework to jointly target both PU classification and conditional generation when exposed to extra data, especially out-of-distribution unlabeled data, by exploring the interplay between them: 1) enhancing the performance of PU classifiers with the assistance of a novel Classifier-Noise-Invariant Conditional GAN~(CNI-CGAN) that is robust to noisy labels, 2) leveraging extra data with predicted labels from a PU classifier to help the generation. Theoretically, we prove the optimal condition of CNI-CGAN and experimentally, we conducted extensive evaluations on diverse datasets. 

\keywords{PU Learning \and Robust Generation \and Unlabeled Data.}
\end{abstract}
\section{Introduction}

Existing machine learning methods, particularly deep learning models, typically require big data to pursue remarkable performance. For instance, conditional deep generative models can generate high-fidelity and diverse images, but they have to rely on vast amounts of labeled data~\cite{lucic2019high}. Nevertheless, collecting large-scale, accurate class-labeled data in real-world scenarios is often laborious or impractical; thus, label scarcity is ubiquitous. Under such circumstances, classification performance and conditional generation~\cite{mirza2014conditional} drops significantly~\cite{lucic2019high}. At the same time, diverse unlabeled data are available in enormous quantities, and therefore, a key issue is how to take advantage of the extra data to enhance the conditional generation or classification. 

For the unlabeled data, both in-distribution and out-of-distribution data exist, where out-of-distribution data does not conform to the distribution of the labeled data. We aim at harnessing the out-of-distribution data to help classification and conditional generation simultaneously. In the generation with extra data, most related works focused on the in-distribution data~\cite{lucic2019high}. For the out-of-distribution data, most existing methods~\cite{yamaguchi2019effective,zhao2020leveraging}  attempted to forcibly train generative models on a large amount of unlabeled data, and then transferred the learned knowledge of the pre-trained generator to the in-distribution data. In classification, a common setting to utilize unlabeled data is semi-supervised learning~\cite{miyato2018virtual,sun2024patch,berthelot2019mixmatch}, but it often assumes the same distribution between the unlabeled and labeled data, ignoring their distributional mismatch. In contrast, Positive and Unlabeled~(PU) Learning~\cite{bekker2020learning,kiryo2017positive} is an elegant way of handling this under-studied problem, where a model has access to both positive samples and unlabeled data.

However, even with the assistance of PU learning to make predictions, it still needs to be determined how to devise a robust conditional generative models against the noisy predicted pseudo labels, posing multiple challenges to investigate the interplay between classification and conditional generation in presence of unlabeled data. Previous work~\cite{hou2017generative} leveraged GANs to recover both positive and negative data distribution to step away from overfitting of PU classifiers, but they never considered the noise-invariant generation or their mutual improvement. The generative-discriminative complementary learning~\cite{xu2019generative} was studied in weakly supervised learning, while we aim to tackle the (Multi-)~Positive and Unlabeled learning, developing the method of noise-invariant generation from noisy labels. 

In this paper, we focus on the mutual benefits of conditional generation and PU classification when we are only accessible to little class-labeled data, but extra unlabeled data, including out-of-distribution data, can be available. Firstly, a parallel non-negative multi-class PU estimator is derived to classify both the positive data of all classes and the negative data. Then we design a \textit{Classifier-Noise-Invariant Conditional Generative Adversarial Network~(CNI-CGAN)} that can learn the clean data distribution on all unlabeled data with noisy labels provided by the PU classifier. Simultaneously, we leverage our CNI-CGAN to re-train the PU classifier to enhance its performance through data augmentation, demonstrating a reciprocal benefit for both generation and classification. We provide the theoretical analysis of the optimal condition for our CNI-CGAN and conduct extensive experiments to verify the superiority of our approach.

\section{Our Method}


\subsection{Positive-Unlabeled Learning}\label{sec:PU}	

\noindent \textbf{Traditional Binary Positive-Unlabeled Problem Setting.} Let $X\in\mathbb{R}^d$ and $Y\in\{\pm1\}$ be the input and output variables and $p(x, y)$ is the joint distribution with marginal distribution $p_p(x)=p(x|Y=+1)$ and $p_n(x)=p(x|Y=-1)$. In particular, we denote $p(x)$ as the distribution of unlabeled data. $n_p$, $n_n$ and $n_u$ are the amount of positive, negative and unlabeled data, respectively. Like most PU learning methods, our method also makes the mild Selected Completely At Random~(SCAR) assumption~\cite{bekker2020learning}, where the labeled samples are selected completely random from the positive distribution.

\noindent \textbf{Parallel Non-Negative PU Estimator.} Vanilla PU learning~\cite{bekker2020learning,kiryo2017positive,du2014analysis,du2015convex} employs unbiased and consistent estimator. Denote $g_{\theta}: \mathbb{R}^{d} \rightarrow \mathbb{R}$ as the score function parameterized by $\theta$, and $\ell: \mathbb{R} \times\{\pm 1\} \rightarrow \mathbb{R}$ as the loss function. The risk of $g_\theta$ can be approximated by its empirical version denoted as $\widehat{R}_{\mathrm{pn}}(g_\theta)$:	$\widehat{R}_{\mathrm{pn}}(g_\theta)=\pi_{\mathrm{p}} \widehat{R}_{\mathrm{p}}^{+}(g_\theta)+\pi_{\mathrm{n}} \widehat{R}_{\mathrm{n}}^{-}(g_\theta)$, where the probability $\pi_p=P(Y=+1)$ with $\pi_p+\pi_n=1$ and $\pi_p$ represents the class prior probability. Denote $\widehat{R}_{\mathrm{p}}^{+}(g_\theta)=\frac{1}{n_p} \sum_{i=1}^{n_{\mathrm{p}}} \ell\left(g_\theta \left(x_{i}^{\mathrm{p}}\right),+1\right)$ and $\widehat{R}_{\mathrm{n}}^{-}(g_\theta)=\frac{1}{n_n} \sum_{i=1}^{n_{\mathrm{n}}} \ell\left(g_\theta \left(x_{i}^{\mathrm{n}}\right),-1\right)$. As negative data $x^n$ are unavailable, a common way is to offset $R_{\mathrm{n}}^{-}(g_\theta)$. We also know that $\pi_{\mathrm{n}} p_{\mathrm{n}}(x)=p(x)-\pi_{\mathrm{p}} p_{\mathrm{p}}(x)$, and hence $\pi_{\mathrm{n}} \widehat{R}_{\mathrm{n}}^{-}(g_\theta)=\widehat{R}_{\mathrm{u}}^{-}(g_\theta)-\pi_{\mathrm{p}} \widehat{R}_{\mathrm{p}}^{-}(g_\theta)$. The resulting unbiased risk estimator $	\widehat{R}_{\mathrm{pu}}(g_\theta)$ is formulated as:	$\widehat{R}_{\mathrm{pu}}(g_\theta)=\pi_{\mathrm{p}} \widehat{R}_{\mathrm{p}}^{+}(g_\theta)-\pi_{\mathrm{p}} \widehat{R}_{\mathrm{p}}^{-}(g_\theta)+\widehat{R}_{\mathrm{u}}^{-}(g_\theta)$, where $\widehat{R}_{\mathrm{p}}^{-}(g_\theta)=\frac{1}{n_p} \sum_{i=1}^{n_{\mathrm{p}}} \ell\left(g_\theta \left(x_{i}^{\mathrm{p}}\right),-1\right)$ and $\widehat{R}_{\mathrm{u}}^{-}(g_\theta)=\frac{1}{n_u} \sum_{i=1}^{n_{\mathrm{u}}} \ell\left(g_\theta \left(x_{i}^{\mathrm{u}}\right),-1\right)$. The advantage of this unbiased risk minimizer is that the optimal solution is accessible if $g$ is linear in $\theta$. However, in practice, more flexible models $g_\theta$, e.g., deep neural networks, are preferred, push the estimator to suffer from overfitting. Hence, we decide to utilize \textit{non-negative risk}~\cite{kiryo2017positive} for our PU learning, which has been verified in \cite{kiryo2017positive} to allow deep neural networks to mitigate overfitting. The non-negative PU estimator is formulated as: $\widehat{R}_{\mathrm{pu}}(g_\theta)=\pi_{\mathrm{p}} \widehat{R}_{\mathrm{p}}^{+}(g_\theta)+\max \left\{0, \widehat{R}_{\mathrm{u}}^{-}(g_\theta)-\pi_{\mathrm{p}} \widehat{R}_{\mathrm{p}}^{-}(g_\theta)\right\}$. We replace $\max \left\{0, \widehat{R}_{\mathrm{u}}^{-}(g_\theta)-\pi_{\mathrm{p}} \widehat{R}_{\mathrm{p}}^{-}(g_\theta)\right\}$ for a parallel implementation of $\widehat{R}_{\mathrm{pu}}(g_\theta)$, with its lower bound $\frac{1}{N}\sum_{i=1}^{N}\max \left\{0, \widehat{R}_{\mathrm{u}}^{-}(g_\theta; \mathcal{X}^i_u)-\pi_{\mathrm{p}} \widehat{R}_{\mathrm{p}}^{-}(g_\theta; \mathcal{X}^i_p)\right\}$, where $\mathcal{X}^i_u$ and $\mathcal{X}^i_p$ denote as the unlabeled and positive data in the $i$-th mini-batch, and $N$ is the number of batches.

\noindent \textbf{From Binary PU to Multi-PU Learning.} Previous PU learning focuses on learning a classifier from positive and unlabeled data and cannot easily be adapted to $K+1$ multi-classification tasks where $K$ represents the number of classes in the positive data. Multi-positive and Unlabeled learning~\cite{xu2017multi} was ever developed, but the proposed algorithm may not allow deep neural networks. Instead, we extend binary PU  learning to the multi-class version in a straightforward way by additionally incorporating cross-entropy loss on all the positive data with labels for different classes. More precisely, we consider the $K+1$-class classifier $f_\theta$  as a score function $f_{\theta} = \left(f_{\theta}^{1}(x), \ldots, f_{\theta}^{K+1}(x)\right)$. After the \textit{softmax} function, we select the first $K$ positive data to construct cross-entropy loss $\ell^{\mathrm{CE}}$, i.e., $\ell^{\mathrm{CE}}(f_\theta(x), y)=\log \sum_{j=1}^{K+1} \exp \left(f_{\theta}^{j}(x)\right)-f_{\theta}^{y}(x)$ where $y\in [K]$. For the PU loss, we consider the composite function $h(f_\theta(x)): \mathbb{R}^{d} \rightarrow \mathbb{R}$ where $h(\cdot)$ conducts a logit transformation on the accumulative probability for the first $K$ classes, i.e., $h(f_\theta(x))=\text{ln}(\frac{p}{1-p})$ in which $p=\sum_{j=1}^{K} \exp \left(f_{\theta}^{j}(x)\right) / \sum_{j=1}^{K+1} \exp \left(f_{\theta}^{j}(x)\right)$. The final mini-batch risk of our PU learning can be presented as:	$\widetilde{R}_{\mathrm{pu}}(f_\theta; \mathcal{X}^i) = \pi_{\mathrm{p}} \widehat{R}_{\mathrm{p}}^{+}(h(f_\theta); \mathcal{X}^i_p) + \widehat{R}_{\mathrm{p}}^{\mathrm{CE}}(f_\theta; \mathcal{X}^i_p)  +  \max \left\{0, \widehat{R}_{\mathrm{u}}^{-}(h(f_\theta); \mathcal{X}^i_u)-\pi_{\mathrm{p}} \widehat{R}_{\mathrm{p}}^{-}(h(f_\theta); \mathcal{X}^i_p)\right\} $, where $\widehat{R}_{\mathrm{p}}^{\mathrm{CE}}(f_\theta; \mathcal{X}^i_p)=\frac{1}{n_p} \sum_{i=1}^{n_{\mathrm{p}}} \ell^{\mathrm{CE}} \left(f_\theta \left(x_{i}^{\mathrm{p}}\right),y\right)$.

\subsection{CNI-CGAN}\label{sec:CNI-CGAN}

\begin{wrapfigure}[13]{r}{0.4\textwidth}
		\vspace{-12mm}
	\includegraphics[width=0.38\textwidth,trim=240 150 250 105,clip]{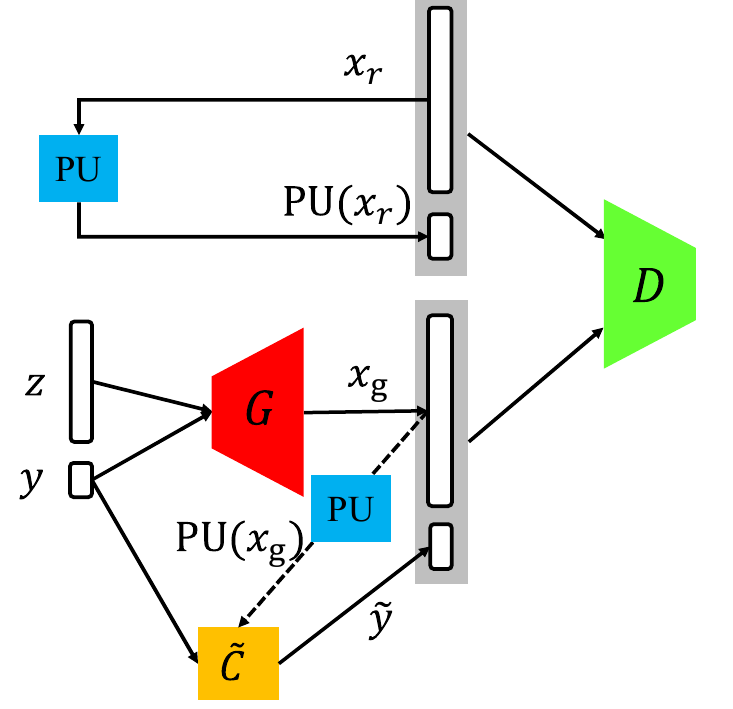}
	\caption{Model architecture of our CNI-CGAN. }
\label{fig:architecture}
\end{wrapfigure}

To leverage extra data, i.e., all unlabeled data, to benefit the generation, we deploy our conditional generative model on all data with pseudo labels predicted by our PU classifier. However, these predicted labels tend to be noisy, reducing the reliability of the supervision signals and thus worsening the performance of the conditional generative model. Besides, the noise depends on the accuracy of the given PU classifier. To address this issue, we focus on developing a novel noise-invariant conditional GAN that is robust to noisy labels provided by a specified classifier, e.g. a PU classifier. We call our method \textit{Classifier-Noise-Invariant Conditional Generative Adversarial Network~(CNI-CGAN)} and the architecture is depicted in Figure~\ref{fig:architecture}.

\noindent \textbf{Principle of the Design of CNI-CGAN.}  Albeit being noisy, the pseudo labels given by the PU classifier still provide rich information that we can exploit. The key is to consider the noise generation mechanism during the generation. We denote the real data as $x_r$ and the predicted hard label through the PU classifier as $PU_{\theta}(x_r)$, i.e., $PU_{\theta}(x_r)=\arg \max_i f^i_\theta(x_r)$, as displayed in Figure~\ref{fig:architecture}.  We let the generator  ``imitate'' the noise generation mechanism to generate pseudo labels for the labeled data. With both pseudo and real labels, we can leverage the PU classifier $f_\theta$ to estimate a confusion matrix $\tilde{C}$ to model the label noise from the classifier. During the generation, a real label $y$, while being fed into the generator $G$, will also be polluted by $\tilde{C}$ to compute a noisy label $\tilde{y}$, which then will be combined with the generated fake sample $x_g$ for the following discrimination. Finally, the discriminator $D$ will distinguish the real samples $[x_r, PU_\theta(x_r)]$ out of fake samples $[x_g, \tilde{y}]$. Overall, the noise ``generation'' mechanism from both sides is balanced.

\noindent \textbf{Estimation of $\tilde{C}$.} The key in the design of $\tilde{C}$ is to estimate the label noise of the pre-trained PU classifier by considering all the samples of each class. More specifically, the confusion matrix $\tilde{C}$ is $k+1$ by $k+1$ and each entry $\tilde{C}_{ij}$ represents the probability of a generated sample $x_g$, given a label $i$, being classified as class $j$ by the PU classifier. Mathematically, we denote $\tilde{C}_{ij}$ as:	
\begin{eqnarray} \begin{aligned}\label{eq:estimiation}
		\tilde{C}_{ij}=P(PU_{\theta}(x_g)=j|y=i)= \mathbb{E}_z[\mathbb{I}_{\{PU_{\theta}(x_g)=j | y=i\}}],
\end{aligned} \end{eqnarray} 
where $x_g=G(z,y=i)$ and $\mathbb{I}$ is the indicator function. Owing to the stochastic optimization nature when training deep neural networks, we incorporate the estimation of $\tilde{C}$ in the processing of training by \textit{Exponential Moving Average~(EMA) method}. This choice can balance the utilization of information from previous training samples and the updated PU classifier to estimate $\tilde{C}$. We formulate the update of $\tilde{C}^{(l+1)}$ in the $l$-th mini-batch as follows: 	$\tilde{C}^{(l+1)}=\lambda \tilde{C}^{(l)}+ (1-\lambda) \Delta_{\mathcal{X}_l}^{\tilde{C}}$, where $\Delta_{\mathcal{X}_l}^{\tilde{C}}$ denotes the incremental change of $\tilde{C}$ on the current $l$-th mini-batch data $\mathcal{X}_l$ via Eq.~\ref{eq:estimiation}. $\lambda$ is the averaging coefficient in EMA.

\noindent \textbf{Theoretical Guarantee of Clean Data Distribution.} Firstly, we denote $\mathcal{O}(x)$ as the oracle class of sample $x$ from an oracle classifier $\mathcal{O}(\cdot)$. Let $\pi_i, i=1,...,K\!+\!1$, be the class-prior probability of the class $i$ in the multi-positive unlabeled setting. Theorem~\ref{theorem:ours} proves the optimal condition of CNI-CGAN to guarantee the convergence to \textbf{the clean data distribution}. The Proof is provided in Appendix~\ref{appendix_CNICGAN}.

\begin{thm}\label{theorem:ours} (Optimal Condition of CNI-CGAN) 
	Given any PU classifier, let $P^g$ be a probabilistic transition matrix where $P^g_{ij}=P(\mathcal{O}(x_g)=j|y=i)$ indicates the probability of sample $x_g$ with the oracle label $j$ generated by $G$ with the initial label $i$. We assume that the conditional sample space of each class is disjoint with each other, then
	
	\noindent (1) $P^g$ is a \textbf{permutation matrix} if the generator $G$ in CNI-CGAN is optimal, with the permutation, compared with an identity matrix, only happens on rows $\mathbf{r}$ where corresponding $\pi_r, r\in \mathbf{r}$ are equal.
	
	\noindent (2) If $P^g$ is an \textbf{identity matrix} and the generator $G$ in CNI-CGAN is optimal, then $p^r(x,y)=p^g(x,y)$ where $p^r(x,y)$ and $p^g(x,y)$ are the real and the generating joint distribution, respectively. 
\end{thm} 
Note that this optimal condition of CNI-CGAN holds for any PU classifier. The assumption of the disjoint conditional sample space allows tractable theoretical results, although it may be violated in practice, potentially degrading the performance of CNI-CGAN. Despite this gap between theory and practice, briefly speaking, CNI-CGAN can learn the clean data distribution if $P^g$ is an identity matrix. More importantly, the method we elaborate on has already guaranteed $P_g$ as a permutation matrix, where the permutation happens only when the same class-prior probabilities exist. While the resulting $P_g$  is nearly close to an identity, we still need an extra constraint to push $P_g$ to be an identity exactly.  

\noindent \textbf{The Auxiliary Loss.} The optimal $G$ in CNI-CGAN can only guarantee that $p^g(x,y)$ is close to $p^r(x,y)$ as the optimal permutation matrix $P^g$ is close to the identity matrix. Hence in practice, to ensure that we can exactly learn an identity matrix for $P^g$ and thus achieve the clean data distribution, we introduce an auxiliary loss to encourage a larger trace of $P^g$, i.e., $\sum_{i=1}^{K+1}P(\mathcal{O}(x_g)=i)|y=i)$. As $\mathcal{O}(\cdot)$ is intractable, we approximate it by the current PU classifier $PU_{\theta}(x_g)$. Then we obtain the auxiliary loss $\ell_{\mathrm{aux}}$:	
\begin{eqnarray} \begin{aligned}
		\ell_{\mathrm{aux}}(z, y)  =  \max \{\kappa - \frac{1}{K+1}\sum_{i=1}^{K+1}\mathbb{E}_z(\mathbb{I}_{\{ PU_{\theta}(x_g)=i|y=i \}}) , 0 \},
	\end{aligned} \nonumber \end{eqnarray} 
where $\kappa \in (0,1)$ is a hyper-parameter. With the support of auxiliary loss, $P^g$ tends to converge to the identity matrix where CNI-CGAN can learn the clean data distribution even in the presence of noisy labels. 

\noindent \textbf{Comparison with RCGAN~\cite{thekumparampil2018robustness,kaneko2019label}.} The theoretical property of CNI-CGAN has a major advantage over existing Robust CGAN~(RCGAN)~\cite{thekumparampil2018robustness,kaneko2019label}, for which the optimal condition can only be achieved when the label confusion matrix is known \textit{a priori}. Although heuristics can be employed, such as RCGAN-U~\cite{thekumparampil2018robustness}, to handle the unknown label noise setting, these approaches still lack the theoretical guarantee to converge to the clean data distribution.

\subsection{Algorithm}

To guarantee the efficacy of our approach, one implicit and mild assumption is that our PU classifier will not overfit on the training data, while our non-negative estimator helps to ensure that it as explained in the previous Section~\ref{sec:PU}. To clarify the optimization process of CNI-CGAN further, we elaborate on the training steps of $D$ and $G$, respectively.

\noindent \textbf{D-Step:}	We train $D$ on an adversarial loss from both real data and generated $(x_g, \tilde{y})$, where $\tilde{y}$ is corrupted by $\tilde{C}$. $\tilde{C}_y$ denotes the $y$-th row of $\tilde{C}$. We formulate the loss of $D$ as:
\begin{eqnarray} \begin{aligned}
		\max_{D \in \mathcal{F}} &\underset{x \sim p(x)}{\mathbb{E}}[\phi(D(x, PU_{\theta}(x)))] + \underset{z \sim P_Z, y \sim P_{Y} \atop \tilde{y} | y \sim \tilde{C}_{y}}{\mathbb{E}}[\phi(1-D(G(z,y), \tilde{y}))],
\end{aligned} \end{eqnarray} 
where $\mathcal{F}$ is a family of discriminators and $P_Z$ is the distribution of latent space vector $z$, e.g., a Normal distribution. $P_Y$ is a discrete uniform distribution on $[K+1]$ and $\phi$ is the measuring function.

\noindent \textbf{G-Step:}	We train $G$ additionally on the auxiliary loss $\ell_{\mathrm{aux}}(z, y)$ as follows:
\begin{eqnarray} \begin{aligned}
		\min _{G \in \mathcal{G}} \underset{z \sim P_Z, y \sim P_{Y} \atop \tilde{y} | y \sim \tilde{C}_{y}}{\mathbb{E}}\left[\phi(1-D(G(z,y), \tilde{y}))+\beta\ell_{\mathrm{aux}}(z, y) \right],
\end{aligned} \end{eqnarray} 	
where $\beta$ controls the strength of auxiliary loss and $\mathcal{G}$ is a family of generators. In summary, our CNI-CGAN conducts $K+1$ class generation, which can be further leveraged to benefit the $K+1$ PU classification via data augmentation.

\noindent \textbf{Procedure.} Firstly, we train a PU classifier $f_\theta$ on a multi-positive and unlabeled dataset with the parallel non-negative estimator. Then, we train our CNI-CGAN, described in the previous Section~\ref{sec:CNI-CGAN}, on all data with pseudo labels predicted by the pre-trained PU classifier. As CNI-CGAN is robust to noisy labels, we leverage the data generated by CNI-CGAN to conduct data augmentation to improve the PU classifier. Finally, we implement the joint optimization for the training of CNI-CGAN and the data augmentation of the PU classifier.

\section{Experiment}\label{sec:experiments}


We first show the reciprocal performance of robust generation and classification via CNI-CGAN. Next, we suggest our proposal is robust against the initial accuracy of PU classifiers as well as the data distribution of the unlabeled dataset.

\noindent \textbf{Experimental Setup.} We perform our approaches and several baselines on MNIST, Fashion-MNIST, and CIFAR-10. We select the first five classes on MNIST and five non-clothes classes on Fashion-MNIST, respectively, for $K+1$ classification~($K=5$). To verify the consistent effectiveness of our method in the standard binary PU setting, we pick the four categories of transportation tools in CIFAR-10 as the one-class positive dataset. As for the baselines, the first is \textbf{CGAN-P}, where a Vanilla CGAN~\cite{mirza2014conditional} is trained only on limited positive data. Another natural baseline is \textbf{CGAN-A} where a Vanilla CGAN is trained on all data with labels given by the PU classifier. The last baseline is \textbf{RCGAN-U}~\cite{thekumparampil2018robustness}, where the confusion matrix is learnable while training. For fair comparisons, we choose the same GAN architecture. Through a line search of hyper-parameters, we choose $\kappa$ as 0.75, $\beta$ as 5.0, and $\lambda=0.99$ across all the datasets.

\noindent \textbf{Evaluation Metrics.} For MNIST and Fashion-MNIST, we mainly use \textbf{\textit{Generator Label Accuracy}}~\cite{thekumparampil2018robustness} and \textbf{\textit{Increased PU Accuracy}} to evaluate the quality of generated images. Generator Label Accuracy compares specified $y$ from CGANs to the true class of the generated examples through a pre-trained (almost) oracle classifier $f$. In experiments, we pre-trained two $K\!+\!1$ classifiers with 99.28$\%$ and 98.23$\%$ accuracy on the two datasets, respectively. Additionally, the increased PU Accuracy measures the closeness between generated data distribution and test (almost real) data distribution for the PU classification, serving as a key indicator to reflect the quality of generated images. For CIFAR 10, we use both \textbf{\textit{Inception Score}}~\cite{salimans2016improved} to evaluate the quality of the generated samples and the increased PU Accuracy to quantify the improvement of generated samples on the PU classification.

\begin{wrapfigure}[13]{r}{0.6\textwidth}
\vspace{-11mm}
	\centering
	\centering\includegraphics[width=0.6\textwidth,trim=10 10 10 0,clip]{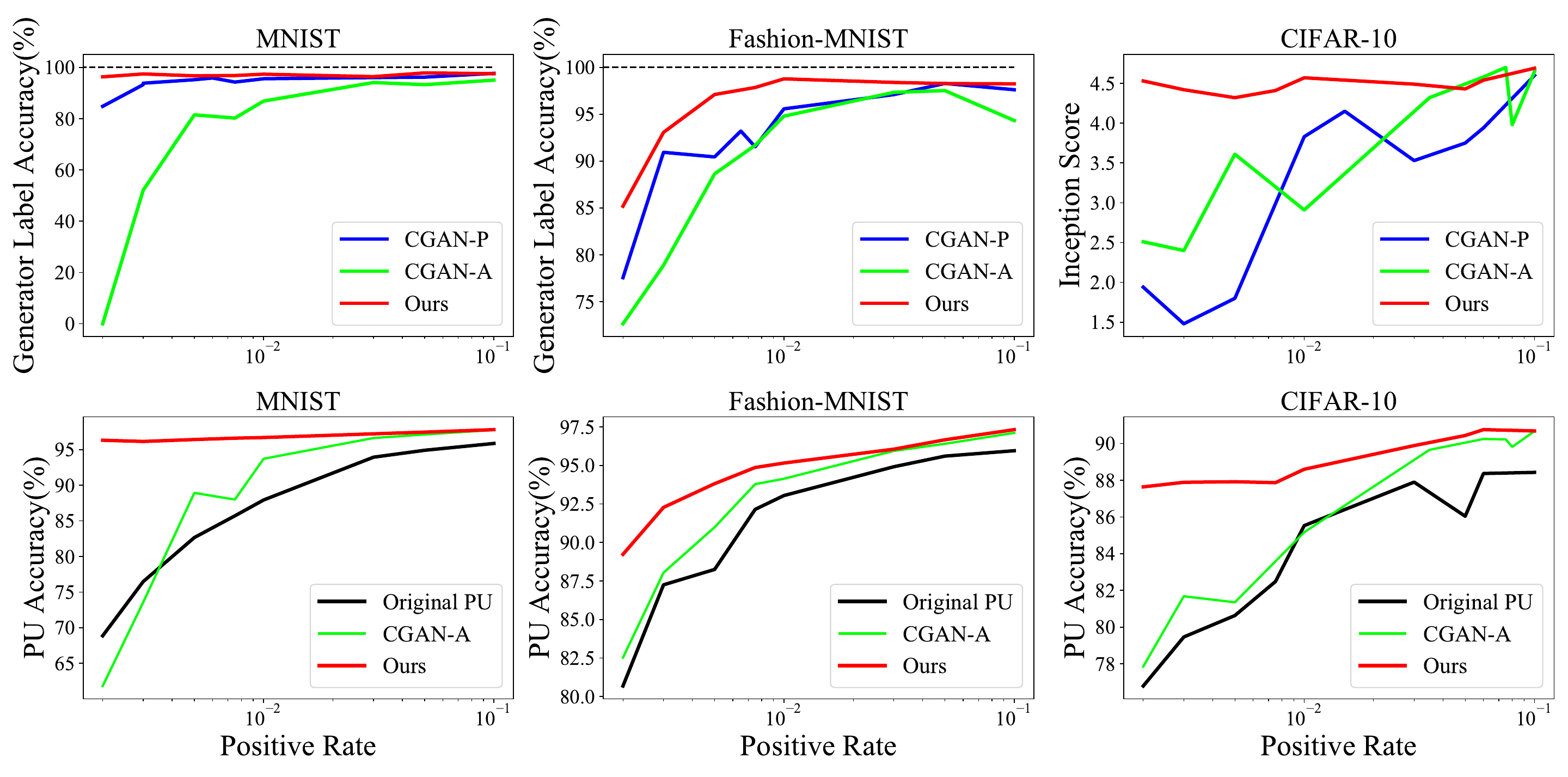}
	\vspace{-8mm}
	\caption{Generation and classification performance of CGAN-P, CGAN-A, and Ours on three datasets. Results of CGAN-P~(blue lines) on PU accuracy do not exist since CGAN-P generates only $K$ rather than $K+1$  class.}
	\label{fig:classification}
\end{wrapfigure}

\subsection{Generation and Classification}
We set the whole training dataset as the unlabeled data and select certain amount of positive data with the ratio of \textit{Positive Rate}. Figure~\ref{fig:classification} presents the trend of Generator Label Accuracy, Inception Score, and PU Accuracy as the Positive Rate increases. It showcases that CNI-CGAN outperforms CGAN-P and CGAN-A consistently, especially when the positive rate is small, i.e., there is little positive data. Remarkably, our approach enhances the PU accuracy greatly when exposed to low positive rates, while CGAN-A even worsens the original PU classifier sometimes in this scenario due to the existence of too much label noise given by a less accurate PU classifier. Meanwhile, when more supervised positive data are given, the PU classifier generalizes better and then provides more accurate labels, conversely leading to more consistent and better performance for all methods. While CGAN-P achieves comparable generator label accuracy on MNIST, it results in a lower Inception Score.


To verify the advantage of theoretical property for our CNI-CGAN, we further compare it with RCGCN-U~\cite{thekumparampil2018robustness,kaneko2019label}, the heuristic version of robust generation against unknown noisy labels setting without the theoretical guarantee of optimal condition. As observed in Table~\ref{table_RCGANU}, our method outperforms RCGAN-U especially when the positive rate is low. When the amount of positive labeled data is relatively large, e.g., 10.0\%, both our approach and RCGAN-U can obtain comparable performance.

\begin{wraptable}[11]{r}{0.6\textwidth}
	\centering
	\vspace{-11mm}
	\caption{PU classification accuracy of RCGAN-U and Ours across three datasets. Final PU accuracy represents the accuracy of PU classifier after the data augmentation.}
	\scalebox{0.6}{
		\begin{tabular}{ccccccc}
			\hline
			\hline
			\multicolumn{2}{c}{\textbf{Final PU Accuracy $\backslash$ Positive Rates~(\%)}}&0.2\%&0.5\%&1.0\%&10.0\%\\
			\hline
			\multirow{3}*{MNIST}&Original PU&68.86&76.75&86.94&95.88\\
			~&RCGAN-U&87.95 &95.24&95.86 &97.80\\
			~                   &Ours   &\bf96.33 &\bf96.43 &\bf96.71&\bf97.82 \\
			\hline	
			\multirow{3}*{Fashion-MNIST}&Original PU&80.68&88.25&93.05&95.99\\
			~&RCGAN-U &89.21   &92.05   &94.59 &97.24\\
			~                   &Ours   &\bf89.23 &\bf93.82 &\bf95.16&\bf97.33 \\
			\hline	
			\multirow{3}*{CIFAR-10}&Original PU&76.79&80.63&85.53&88.43\\
			~&RCGAN-U &83.13  &86.22   &88.22   &90.45\\
			~ &Ours   &\bf87.64&\bf87.92 &\bf88.60&\bf90.69 \\
			\hline
			\hline
		\end{tabular}
		
	}
	\label{table_RCGANU}
\end{wraptable}

\begin{figure*}[htbp]
	\centering
		\vspace{-5mm}
	\centering\includegraphics[width=0.9\textwidth,trim=10 180 30 180,clip]{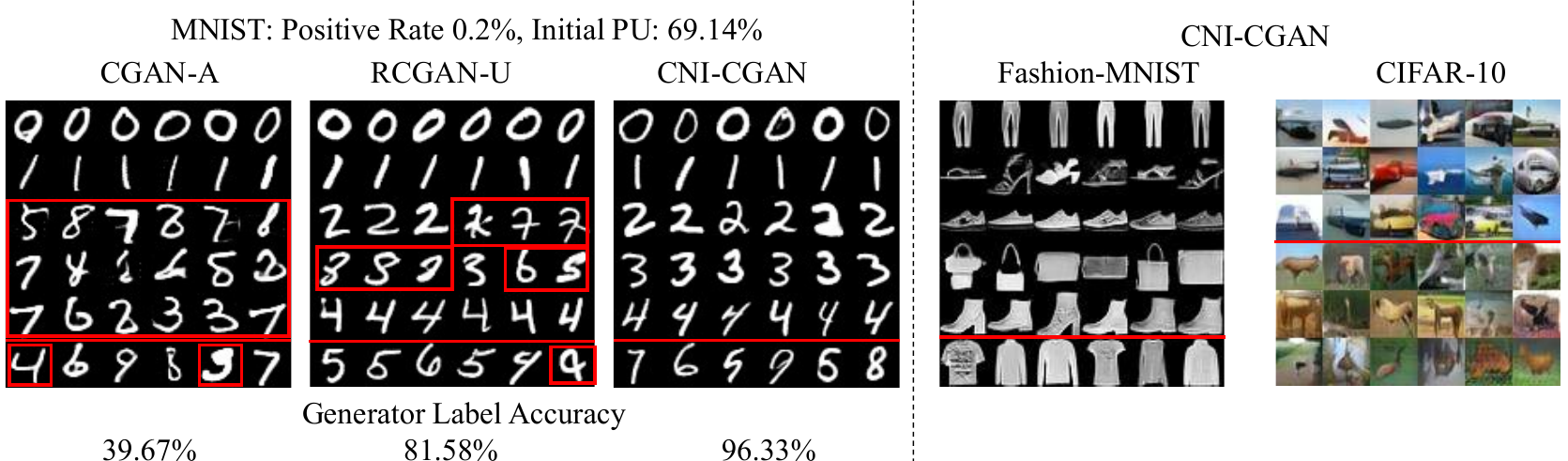}
		\vspace{-4mm}
	\caption{Visualization of generated samples on three datasets. The rows below the red line represent the negative class. We highlight the erroneously generated images with red boxes on MNIST.}
	\label{fig:visualization}
\end{figure*}

\noindent \textbf{Visualization.} To further demonstrate the superiority of CNI-CGAN compared with the other baselines, we present some generated images within $K+1$ classes from CGAN-A, RCGAN-U, and CNI-CGAN on MNIST, and high-quality images from CNI-CGAN on Fashion-MNIST and CIFAR-10, in Figure~\ref{fig:visualization}. In particular, we choose the positive rate as 0.2$\%$ on MNIST, yielding the initial PU classifier with 69.14$\%$ accuracy. Given the noisy labels on all data, our CNI-CGAN can generate more accurate images of each class visually compared with CGAN-A and RCGAN-U.

\vspace{-11mm}

\subsection{Robustness of Our Approach} 

\begin{wrapfigure}[10]{r}{0.6\textwidth}
	\vspace{-8mm}
	\centering
	\centering\includegraphics[width=0.6\textwidth,trim=10 100 20 50,clip]{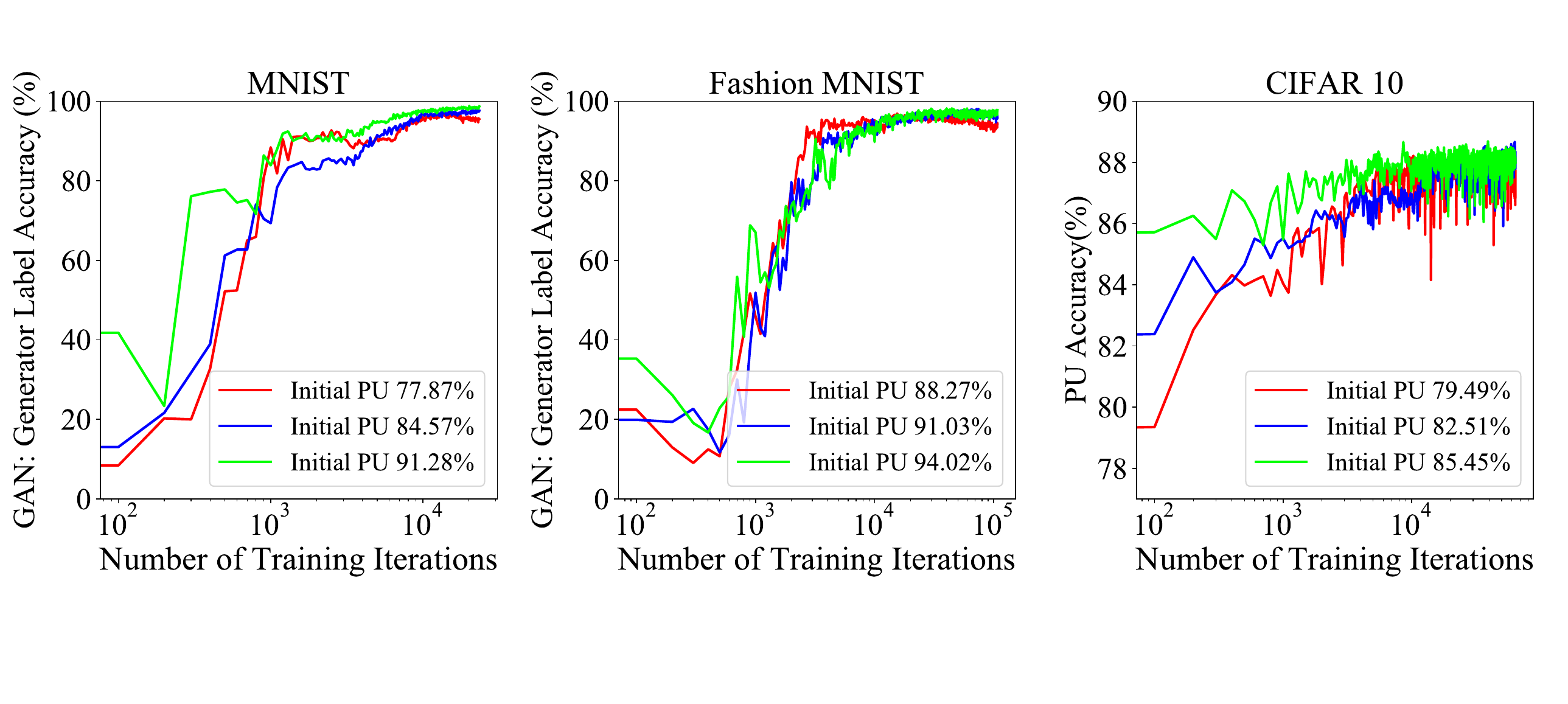}
	\vspace{-7mm}
	\caption{Robustness against the Initial PU accuracy. The tendency of generation performance as the training iterations increases on three datasets.}
	\label{fig:speed}
	
\end{wrapfigure}

We show that our proposed algorithm is robust against the initial accuracy of PU classifiers as well as the typical data distribution of the unlabeled dataset.

\noindent \textbf{Robustness against the Initial PU accuracy.} The auxiliary loss can help the CNI-CGAN to learn the clean data distribution regardless of the initial accuracy of PU classifiers. To verify that, we select distinct positive rates, yielding the pre-trained PU classifiers with different initial accuracies. Then, we perform our method based on these PU classifiers. Figure~\ref{fig:speed} suggests that our approach can still attain similar generation quality under different initial PU accuracies after sufficient training, although better initial PU accuracy can be beneficial to the generation performance in the early phase.

\begin{wrapfigure}[13]{r}{0.6\textwidth}
	\vspace{-12mm}
	\centering
	\centering\includegraphics[width=0.6\textwidth,trim=100 20 130 60,clip]{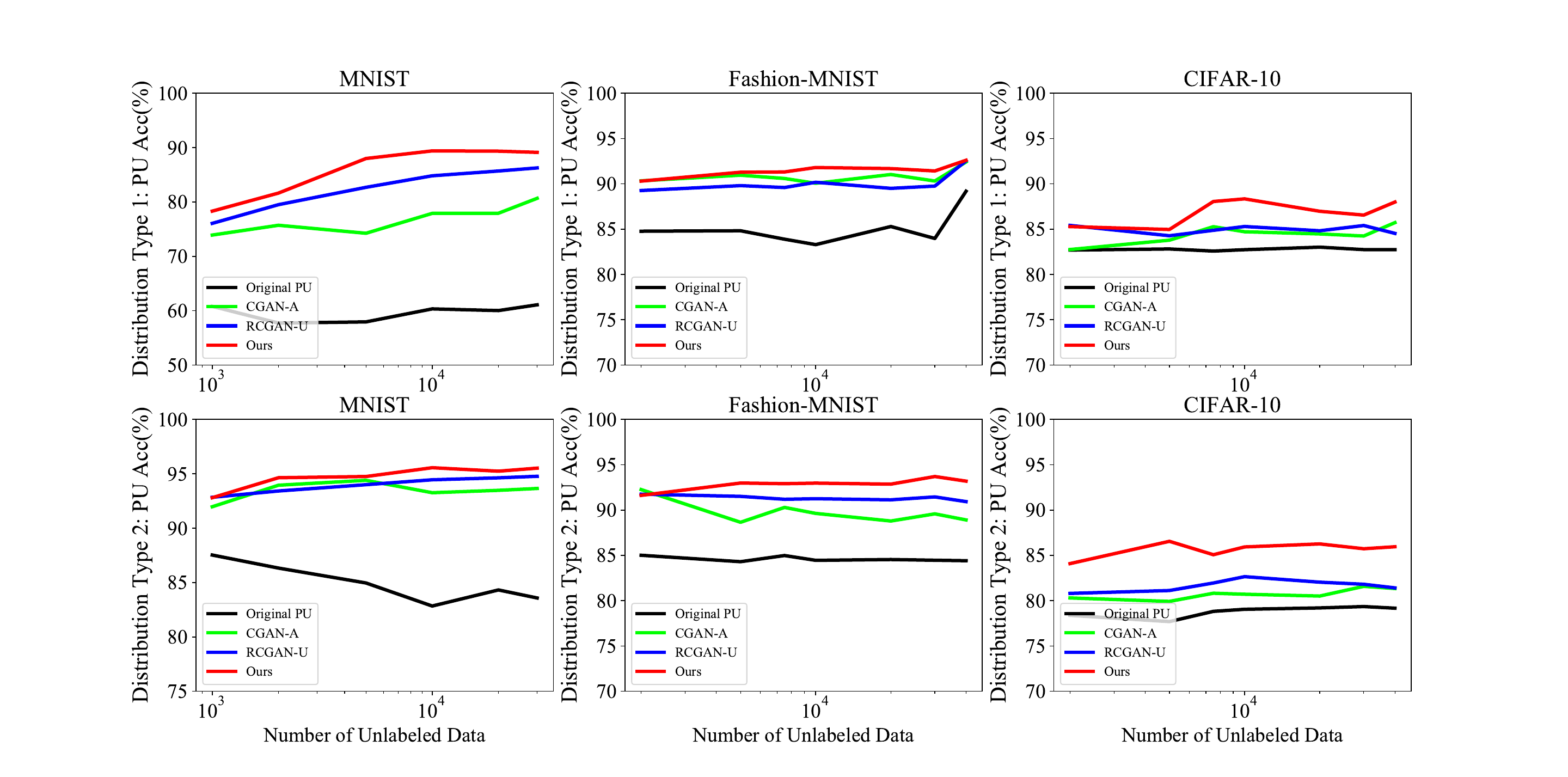}
		\vspace{-8mm}
	\caption{Robustness against the Unlabeled data. PU Classification accuracy of CGAN-A, RCGAN-U, and Ours after joint optimization across different amounts and distribution types of unlabeled data.}
	\label{fig:sensitivity}
\end{wrapfigure}

\noindent \textbf{Robustness against the Unlabeled data.} In real scenarios, we are more likely to have little knowledge about the extra data we have. To further verify the Robustness of CNI-CGAN against the unknown distribution of extra data, we test different approaches across different amounts and distributions of the unlabeled data. Particularly, we consider two different types of distributions for unlabeled data. Type 1 is $[\frac{1}{K+1}, ... ,\frac{1}{K+1},\frac{1}{K+1}]$ where the number of data in each class, including the negative data, is even, while type 2 is $[\frac{1}{2K}, ... \frac{1}{2K},\frac{1}{2}]$ where the negative 
data makes up half of all unlabeled data. In experiments, we focus on PU Accuracy to evaluate both the generation quality and the improvement of PU learning. For MNIST, we choose 1$\%$ and 0.5$\%$ for two settings while we opt for 0.5$\%$ and 0.2$\%$ on both Fashion-MNIST and CIFAR-10. Figure~\ref{fig:sensitivity} manifests that the accuracy of the PU classifier exhibits a slight ascending tendency with the increasing number of unlabeled data. More importantly, our CNI-CGAN almost consistently outperforms other baselines across different amounts of unlabeled data as well as distinct distributions of unlabeled data. This verifies that the Robustness of our proposal for the distribution of extra data can potentially be maintained. We leave the investigation on robustness against more imbalanced situations as future work.

\vspace{-5mm}

\section{Discussion and Conclusion}

\vspace{-5mm}

Due to the computation cost, we leave the efficacy demonstration of our approach on larger datasets, such as ImageNet, as future works. Since our CNI-CGAN approach is agnostic to the estimator choice of PU classification and thus is beneficial to general PU classifiers, we mainly focus on the most typical multi-PU non-negative estimator we elaborated in our work, while leaving the further investigation on more PU estimators as the extension of our work in the future.

In this paper, we proposed CNI-CGAN that breaks the ceiling of class-label scarcity by combining two promising yet separate methodologies to gain massive mutual improvements. CNI-CGAN can learn the clean data distribution from noisy labels given by a PU classifier and then enhance the performance of PU classification through data augmentation in various settings. We have demonstrated, both theoretically and experimentally, the superiority of our proposal on diverse benchmark datasets exhaustively and comprehensively. 

\section*{Acknowledgment}

Z. Lin was supported by the NSF China (No. 62276004).

%
%

\appendix

\section{Proof of Theorem~\ref{theorem:ours}}\label{appendix_CNICGAN}
Firstly, we recall some definitions. Denote $x_r$, $x_g$ as the real training and generated samples, respectively. $x$ are the population of all data, and $x_r$ are sampled from $p(x)$. $y_g$ represents the initial labels for the generator $G$, while $\tilde{y}$ indicates the labels perturbed by $\tilde{C}$ from $y_g$.	The class-prior $\pi_i$ meets $\pi_i=P(y_g=i)=P(\mathcal{O}(x_r)=i)$. For a rigorous proof of Theorem~\ref{theorem:ours}, we elaborate it again in the appendix.

\paragraph{Theorem~\ref{theorem:ours}} We assume that the following three mild assumptions can be met: (a) PU classifier is not overfitting on the training data, (b) $P(PU_\theta(x_g)|\mathcal{O}(x_g), y_g)=P(PU_\theta(x_g)|\mathcal{O}(x_g))$, (c) the conditional sample space is disjoint from each other class. Then,

\noindent (1) $P^g$ is a permutation matrix if the generator $G$ in CNI-CGAN is optimal, with the permutation, compared with an identity matrix, only happens on rows $\mathbf{r}$ where corresponding $\pi_r, r\in \mathbf{r}$ are equal.

\noindent (2) If $P^g$ is an identity matrix and the generator $G$ in CNI-CGAN is optimal, then $p^r(x,y)=p^g(x,y)$ where $p^r(x,y)$ and $p^g(x,y)$ are the real and generating joint distribution, respectively. 

\paragraph{Proof of (1)} For a general setting, the oracle class of $x_g$ given by label $y_g$ is not necessarily equal to $PU_\theta(x_g)$. Thus, we consider the oracle class of $x_g$, i.e., $\mathcal{O}(x_g)$ in the Proof.

\paragraph{Optimal $G$.} In CNI-CGAN, $G$ is optimal if and only if
\begin{equation}\label{eq:RCGAN_G}
	p^r(x_r,PU_\theta(x_r))=p^g(x_g,\tilde{y}).
\end{equation}
The equivalence of joint probability distribution can further derive the equivalence of marginal distribution, i.e., $p^r(x_r)=p^g(x_g)$. We define a probability matrix $C$ where $C_{ij}=P(PU_{\theta}(x)=j|\mathcal{O}(x)=i)$ where $x$ are the population data. According to (c), we can apply $\mathcal{O}(\cdot)$ on both $x_r$ and $x_g$ in Eq.~\ref{eq:RCGAN_G}. Then we have:

\begin{footnotesize}
	\begin{equation}\begin{aligned} 
			P(\mathcal{O}(x_r)=i, PU_\theta(x_r)=j)\overset{(c)}{=}&P(\mathcal{O}(x_g)=i, \tilde{y}=j)\\
			P(\mathcal{O}(x_r)=i) P(PU_{\theta}(x_r)=j|\mathcal{O}(x_r)=i)=& \sum_{k=1}^{K+1}  P(y_g=k, \mathcal{O}(x_g)=i)P(\tilde{y}=j|y_g=k, \mathcal{O}(x_g)=i)\\
			\pi_i C_{ij}\overset{(a)}{=}&\sum_{k=1}^{K+1}  P(\mathcal{O}(x_g)=i|y_g=k)P(y_g=k)P(\tilde{y}=j|y_g=k)\\
			\pi_i C_{ij}=&\sum_{k=1}^{K+1} P^{g \top}_{ik} \pi_k \tilde{C}_{kj},
	\end{aligned} \end{equation}
\end{footnotesize}

where assumption (a) indicates that $PU_\theta(x_r)$ is close to $PU_\theta(x)$ so that $P(PU_{\theta}(x_r)=j|\mathcal{O}(x_r)=i)=P(PU_{\theta}(x)=j|\mathcal{O}(x)=i)$. Then the corresponding matrix form follows as
\begin{equation}\begin{aligned} \label{eq:PURCGAN_GAN}
		\Pi C=P^{g \top} \Pi \tilde{C}
\end{aligned} \end{equation}

\paragraph{Definition.}  According to the definition of $\tilde{C}$ and Law of Total Probability, we have:	
\begin{footnotesize}
	\begin{equation}\begin{aligned} 
			P(y_g=i) P(PU_{\theta}(x_g)=j|y_g=i)=&\pi_i\sum_{k=1}^{K+1} P(\mathcal{O}(x_g)=k|y_g=i)P(PU_{\theta}(x_g)=j|\mathcal{O}(x_g)=k,y_g=i)\\
			\pi_i\tilde{C}_{ij}\overset{(b)}{=}&\pi_i\sum_{k=1}^{K+1} P^g_{ik} P(PU_{\theta}(x_g)=j|\mathcal{O}(x_g)=k)\\
			\pi_i\tilde{C}_{ij}=&\pi_i\sum_{k=1}^{K+1} P^g_{ik} C_{kj},
	\end{aligned} \end{equation}
\end{footnotesize}
where the last equation is met as $p(x_g)$ is close to $p(x)$ when $G$ is optimal, and thus $P(PU_{\theta}(x_g)=j|\mathcal{O}(x_g)=k)=P(PU_{\theta}(x)=j|\mathcal{O}(x)=k)$. Then we consider the corresponding matrix form as follows
\begin{equation}\begin{aligned} \label{eq:PURCGAN_Definition}
		\Pi\tilde{C}=\Pi P^g C
\end{aligned} \end{equation}
where $\Pi$ is the diagonal matrix of prior vector $\pi$. Combining Eq.~\ref{eq:PURCGAN_Definition} and \ref{eq:PURCGAN_GAN}, we have $P^{g \top} \Pi P^{g}=\Pi$, which indicates $P^{g}$ is a general orthogonal matrix. In addition, the element of $P^{g}$ is non-negative and the sum of each row is 1. Therefore, we have $P^g$ as a permutation matrix with permutation compared with the identity matrix only happens on rows $\mathbf{r}$ where corresponding $\pi_r, r\in \mathbf{r}$ are equal. Particularly, if all $\pi_i$ are different from each other, then permutation operation will not happen, indicating the optimal conditional of $P^g$ is the identity matrix.

\paragraph{Proof of (2)}
We additionally denote $y_r$ as the real label of real sample $x_r$, i.e., $y_r=\mathcal{O}(x_r)$. According to the optimal condition of $G$ in Eq.~\ref{eq:RCGAN_G}, we have $p^r(x_r)=p^g(x_g)$. Since we have $P^g$ is an identity matrix, then $\mathcal{O}(x_g)=y_g$ a.e. Thus, we have $p^g(x_g|y_g=i)=p^g(x_g|\mathcal{O}(x_g)=i), \forall i=1,..,K+1$. According the assumption (c) and Eq.~\ref{eq:RCGAN_G}, we have $p^r(x_r|\mathcal{O}(x_r)=i)=p^g(x_g|\mathcal{O}(x_g)=i)$. In addition, we know that $p^r(x_r|\mathcal{O}(x_r)=i)=p^r(x_r|y_r=i)$, thus we have $p^r(x_r|y_r=i)=p^g(x_g|y_g=i)$. Further, we consider the identical class-prior $\pi_i$. Finally, we have 	
\begin{equation}\begin{aligned} 
		p^r(x_r|y_r=i)\pi_i&=p^g(x_g|y_g=i)\pi_i\\
		p^r(x_r|y_r=i)p(\mathcal{O}(x_r)=i)&=p^g(x_g|y_g=i)p(y_g=i)\\
		p^r(x_r|y_r=i)p(y_r=i)&=p^g(x_g|y_g=i)p(y_g=i)\\
		p^r(x_r,y_r)&=p^g(x_g,y_g).
\end{aligned} \end{equation}

 \bibliographystyle{splncs04}
\bibliography{PURCGAN}

\end{document}